# Applying the Negative Selection Algorithm for Merger and Acquisition Target Identification

Theory and Case Study

Satyakama Paul
University of Johannesburg,
Dept. of Mechanical Eng. Science, FEBE, APK Campus,
Johannesburg, 2006, South Africa.
psatyakama@student.uj.ac.za

Andreas Janecek
University of Vienna,
Research Group Theory and Applications of Algorithms
1090 Vienna, Austria.
andreas.janecek@univie.ac.at

Fernando Buarque de Lima Neto
University of Permambuco, POLI (Computing Eng. Program),
RuaBenfica, 455 – Madalena, 50.720-001,
Recife/ PE, Brazil.
fbln@ecomp.poli.br

Tshilidzi Marwala
University of Johannesburg,
Office of the Deputy Vice Chancellor (RTI), APK Campus,
Johannesburg, 2006, South Africa.
tmarwala@uj.ac.za

*Abstract*— In this paper, we propose a new methodology based on the Negative Selection Algorithm that belongs to the field of Computational Intelligence (specifically, Artificial Immune Systems - AIS) to identify takeover targets. Although considerable research based on customary statistical techniques and some contemporary Computational Intelligence techniques have been devoted to identify takeover targets, most of the existing studies are based upon multiple previous mergers and acquisitions. Contrary to previous research, the novelty of this proposal lies in the methodology's ability to suggest takeover targets for novice firms that are at the beginning of their merger and acquisition spree. We first discuss the theoretical perspective and then provide a case study with details for practical implementation, both capitalizing from unique generalization capabilities of AIS algorithms.

I. INTRODUCTION

Merger and Acquisition (M&A) is of crucial importance in today's business. In 2012, the global M&A was of the aggregate deal volume of USD 2.23 trillion, of which the largest share (18%) was held by the financial industry [1]. Although in recent times researchers in financial management have shown significant interests in predicting future takeover targets in M&A scenarios, recent studies[2] indicate that most companies (included in this set are also the large companies) do not have structured systems to support their M&A decisions. This lack of support is further aggravated for those acquiring firms which would like to start their M&A processes.

Obviously, the limited history data of the later could prevent or at least be a serious hindrance for successful takeovers. This is because the traditional research of M&A base their predictions of the next takeover target by analyzing the statistical records of firms with multiple previous M&A experiences.

The present research is targeted towards acquiring firms which have no previous history of M&A. Our objective is to propose a new methodology that can be easily used and yet with some reasonable assurance be used by new comers in the M&A risky business. The approach applies Negative Selection Algorithm (NSA) from Artificial Immune System (AIS) in order to support such novice firms that would like to carry out their first acquisition. Our motivation in using NSA lies in its capability of distinguishing between the *self* and the *nonself*, when only the *self* is available [3]. So differently from other adaptive methods that would require lots of previous data, for example Artificial Neural Networks (ANN); AIS and specifically NSA can produce adequate generalization in the absence of many examples. This most interesting feature is directly drawn from immune systems of vertebrates in which the number of solutions available (i.e. antibodies) are much less compared to the diseases, so nature has evolved means to vary the few existing examples of what are *self* cells to distinguish from target *nonself* cells (i.e. pathogens). And to verify our theoretical proposition we carried out a case study from real data related to the Indian banking industry.

The structure of the paper is as follows: in Section II, we provide a review of the previous works in the area; in Section III, we give a brief biological explanation of the vertebrate immune system; in Section IV, we draw inspiration from NSA and AIS to customize the NSA's application in the present challenging and contemporary problem; in Section V, we examine how NSA can be applied to predict the takeover target of a banking firm that would carry out its first acquisition; and finally, in Section VI, we conclude our work and produce some comments for further research.

## II. REVIEW OF LITERATURE

So far there have been a large number of statistical techniques have been successfully used for predicting M&A takeover targets. Examples of such techniques are univariate analysis [4]; multivariate discriminant analysis [5] [6]; and various forms of regression analysis such as probit/logit [7] [8] [9] [10] [11], nominal logistic regression [12], multinomial logit [13], binomial and multinomial logit [14], and logit [15]. Barnes [16] states that the choice and appropriateness of the statistical techniques are dependent on the statistical nature of the data. He explains that compared to discriminant analysis (DA) which stipulated the conditionality that the "explanatory variables are jointly normal with equal covariance matrix" [16]; logit is more unrestrictive with no such conditionality on the explanatory variables. More recently, Liu and Hu [17] suggested that the statistical techniques predict the future takeover target by matching the characteristics of the previously acquired firms with the characteristics of the potential future takeover targets. However, the serious limitation of the parametric statistical techniques is that they do not adapt themselves in line with the changing nature of M&A problem.

More recently, Computational Intelligence (CI) techniques such as Artificial Neural Network (ANN) and their variations such as Self Organizing Maps (SOMs) of Hopfield Neural Networks (HNNs) have been used for predicting takeover targets. In arguing for the benefit of ANN; Cheh, Weinberg and Yook[18] claim that the parametric nature of the statistical techniques requires one to make certain assumptions about the exact nature of the functional relationship between the multiple input variables. However, such a problem can be avoided by using ANNs which do not need any functional relationship between the multiple input variables. In order to identify the potential takeover targets, they have used a standard feed-forward ANN with back propagation and one hidden layer and compare their results with DA for the same data. In their study, they show that the type I error from the ANN is lower than the type I error of DA, which indicates that compared to ANN, DA is more likely to wrongly classify companies as acquired, when they are actually not. The type II error of the DA is 0.4355, compared to the type II error of ANN (0.5860 – 0.8763), which indicates that compared to ANN, DA is better at classifying companies as not acquired when in practice they are acquired [19]. Hongjiu, Yanrong, and Shufen [20] and Little, Hickey, and Brabazon [21] distinguish between target and non-target companies by using SOM. Although the study in [20] does not provide an accurate value of their classification accuracy, the authors claim that their results are in accordance with the expert judgment of M&A target selection. In [21], the authors achieve an in-sample accuracy of 94.80% for 100 companies that are successfully acquired; and an out-of-sample accuracy of 95.20% for 100 companies that are not acquired. They also claim that SOM had the capability of identifying acquisition targets with greater than 92% accuracy with data of one year prior to the acquisition. Lastly Liu & Hu [17] use HNN to identify potential M&A targets. First they use multiple histories of successful M&As to train the HNN and create an index system for target recognition. Subsequently the trained HNN predict two companies as takeover targets by comparing the index system and 21 potential takeover target firms. Among the two identified takeover targets, one ultimately turns out to be an actual acquisition (regardless of the HNN predictions and is based upon other managerial considerations).

Also an examination of the experimental set-up of the most of the above reviewed studies shows that the researchers consider multiple histories of previous M&A to predict the next takeover target. Thus while Stevens [5] consider 45 acquisition, Barnes [6] work with 13 acquisitions. Harris, Stewart, Guilkey, and Carleton's [7] analysis is based upon a sample of 61 and 45 acquired companies. Meador, Church, and Rayburn's [9] use a sample of 100 acquired companies; and Tsagkanos, Georgopoulos, and Sirioupoulos [10] analysis consist of 35 acquired companies. Cheh, Weinberg and Yook [18] use a sample space of 1275 unacquired companies and 173 acquired companies. Hongjiu, Yanrong, and Shufen [20] use a set of ten enterprises to distinguish between target and non-target companies. Little, Hickey, and Brabazon [21] use a total of 200 quoted US to separate and cluster the merged and non-merger company. Lastly, Liu and Hu [17] use 21 acquired companies to predict the future takeover targets. Compared to the above research, we deviate in our approach of takeover target prediction by considering a novice firm that has no previous experience of M&A. Thus the novelty of our approach lies in predicting takeover targets without any need of any sample space of previous acquisitions.

## III. BIOLOGICAL PERSPECTIVE

The vertebrate immune system consists of a complex multi-layered system of diverse organs, tissues, cells and molecules to combat against the harmful infectious and invading agents (e.g. bacteria, virus, etc), called pathogens. The immune system responds to the pathogens in two ways: (i) through a less specific component called innate immunity (that is non-adaptive), and (ii) a more specific component called adaptive immunity. The body provides an immune response primarily through the leukocytes, the most important ones being phagocytes and lymphocytes [22]. As the first line of defense in the innate immune system, the phagocytes engulf the pathogen, internalize and destroy them. However these phagocyte cells are non-specific in their response. The lymphocytes act as the second line of defense and are adaptive in their immune response. The two important lymphocytes relevant to the present discussion are the B lymphocytes (B-cells) and T lymphocytes (T-cells). When antigen receptors (antibodies) of the T-cells match and bind with antigen, the activated T-cell releases lymphokines that subsequently activate the B-cell [23]. The activated B-cells then produce antibodies to destroy the antigens. However before the B-cells produce antibodies, they undergo somatic mutation to introduce diversity into the B-cell population and then, successful ones are cloned in large numbers [24] [3]. Based upon such inspiration of the natural immune system, AIS is defined as an "Adaptive systems, inspired by theoretical immunology and observed immune functions, principles and models, which are applied to problem solving." [25]. It may be worthwhile to mention here that AIS have been successfully used in the past to study diverse areas of business. Examples of such areas include - its application in supply chain transportation problems [26]; market segmentation [27], financial pattern recognition and identification of financial at-risk companies [28], learning [29], intrusion detection in 'wi-fi' networks [30] etc. Although AIS encompasses several different types of algorithms, the most important ones are based upon Danger Theory, Clonal Selection, Immune Networks, and Negative Selection [25].

While a detailed theoretical explanation of all AIS algorithms is beyond the scope of this research, we discuss the NSA. There exist possibilities that the lymphocytes that react with the pathogens to destroy them, can also react with the host body's own cells (*self* cells) attacking them. Pointed out by Joshua Lederberg in 1959, Negative Selection provides a mechanism to protect the *self* cells of the host, and also destroy unknown antigens (*nonself* cells). Within the highly-impregnable barrier of thymus, the thymocytes (immature T-cells) mature by a pseudo-random genetic rearrangement of its receptors. Next (still in the thymus) these mature T-cells are exposed to *self*-peptides, and those that react strongly with the peptides are eliminated through a process called apoptosis. The rest of the mature T-cells that do not react with the *self*-peptides are subsequently released outside the thymus, into the body, to fight against *nonself* antigens. The result of such a mechanism is that while on the one hand the (released) matured T-cells kill the *nonself* antigens; they are, on the other hand, non-reactive to the *self* (body) cells. Thus NSA may be viewed as a mechanism to discriminate the *self* from the *nonself* [3], mechanism that is seminal in this proposed application.

## IV. ADAPTATION OF NSA TO TAKEOVER TARGET PREDICTION

Let us assume that with intension for M&A, an acquiring company **A** (*self*) wants to identify its takeover target company or companies from a finite set of potential (*nonself*) takeover targets **{b, c, d, e, f, g, h}**[1]. In actual practice there are several reasons for M&A, examples of such include: increase of production scale and efficiency; expansion of geographic coverage; incorporation of new and different knowledge, skill set etc; and cross industry association to combine complimentary products and services [31]. However for the sake of simplicity it is assumed here that the *self* and *nonself* companies produce the same product or service. Adopting the NSA to the present problem, the flowchart seen in Fig. 1 of the model would consist of the following steps:

Step 1: Feature selection is an important starting point of the model. For the acquiring company, the choice of a correct set of indicator variables (features) may go a long way in accurately identifying its compatible target. Ideally the feature set should consist of indicator variables from the company's past and present production, financial, marketing, and human resource conditions; and also the industry and other macro-economic conditions [20]. However, there are no hard and fast rules for such choices as they depend on the judgment of the acquiring company's M&A experts and the availability of data of both the acquirer and the potential takeover targets. We assume that one should be able to identify appropriate separate feature variables, we suggest 18 as enumerated in Table 1, the notation used for features is $\{x^1, x^2, \ldots x^{18}\}$. Table 1 provides a list of the suggested features, where feature 1-14 are absolute values, 15 is in percentage, and 16-18 are ratio values.

TABLE I.    LIST OF FEATURES.

| *Features* | | | |
|---|---|---|---|
| 1. | Number of offices | 10. | Other income |
| 2. | Number of employees | 11. | Interest expended |
| 3. | Business per employee | 12. | Operating expenses |
| 4. | Profit per employee | 13. | Cost of funds (CoF) |
| 5. | Capital and Reserves & Surplus | 14. | Return on Assets |
| 6. | Deposits | 15. | Wages as % to total expenses |
| 7. | Investments | 16. | Return on Advances adjusted to CoF |
| 8. | Advances | 17. | Capital-to-Risk-Weighted Assets Ratio |
| 9. | Interest income | 18. | Net NPA Ratio |

Step 2: Since most M&A decisions are based upon the performance of the acquiring company over a considerable period of time; consider, that our acquiring company **A** reaches its decision based on historical data over a period of many years, $x^i_j$ represents the value of the feature variable **i** in a year **j**. Say for year, the notation would be, **j** = {1, 2, .. 4}. A tabulation of these $x^i_j$ values, for the example of four years, gives us the feature space of the *self* company **A** as in Fig. 2.

---
[1] It is to be noted here that we arbitrarily assume the number of non-*self* companies, (and subsequently) features and years to be: 7, 18 and 4 respectively. This is done to create uniformity and hence an ease of understanding the theoretical perspective and forthcoming case study. However, depending on the use of the practitioner, the model can easily be replicated with any value of non-*self*'s, features and years.

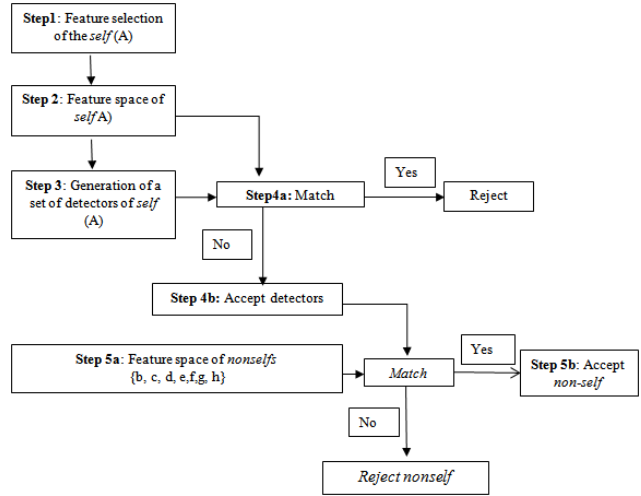

Fig.1. Flowchart of the model.

| $X^1_1$ | $X^2_1$ |  | ... | $X^{17}_1$ | $X^{18}_1$ |
|---|---|---|---|---|---|
| $X^1_2$ | $X^2_2$ |  | ... | $X^{17}_2$ | $X^{18}_2$ |
| $X^1_3$ | $X^2_3$ |  | ... | $X^{17}_3$ | $X^{18}_3$ |
| $X^1_4$ | $X^2_4$ |  | ... | $X^{17}_4$ | $X^{18}_4$ |

Fig.2. Feature space of the *self* (**A**) in the working example.

Step 3 – Detector generation: The next proposed step is to generate the set of detectors of the *self* (**A**). For the applications of AIS mentioned in Section III, the authors used a random process to select the detectors. This is mostly because these areas are highly complex and difficult to understand and predict. Thus, a complete stochastic process of selecting the detectors suits the purpose of the area. However, in a slight deviation from the above application of NSA, we use a combination of deterministic and stochastic processes for selection of detectors. This is justified by the fact that our present system has two components: deterministic component, incorporated into the model in considering the fact that human judgment (e.g. in identifying the span of the detectors) is certainly more appropriate than completely randomized processes; and (ii) stochastic component, included in the model for considering external and internal environmental factors. It is important to notice that such factors affect the functioning and decision making of organizations [32].

A practical approach to detector generation might be to find the average and standard deviation of each of the 18 feature variables, over the period of four years (as suggested in the illustrative example). Let us denote a detector by $dx^i = \{x^1, x^2, ..., x^{18}\}$, where **i** represents a feature. The range of each detector ($dx^i$) can vary from $[(\mu^i_{x_j} - n\delta^i_{x_j} - C^i_j), (\mu^i_{x_j} + n\delta^i_{x_j} + C^i_j)]$, where **μ** and **δ** represents the mean and standard deviation of each of the feature variables, respectively, $C^i_j$ is the change due to changes in external and internal environmental factors. **n** can be viewed as an index of span of the detector and $0 \leq n \leq (\mu^i_{x_j}/\delta^i_{x_j})$. In the following paragraph we clarify the significance of **n**. A schematic diagram of the detector space vector is given below in Fig.3.

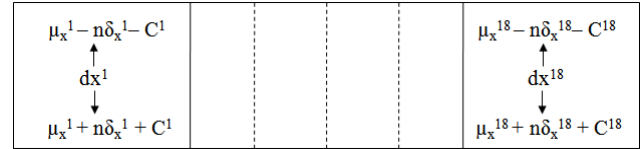

Fig.3. Detector space vector of the *self* (**A**) in the working example.

The rationale behind any merger or acquisition process is to create a combined value that is greater than the sum of the parts [32]. We propose that this can be seen as Value $_{(self + nonself)}$ > Value $_{(self)}$ + Value $_{(nonself)}$, where *self* refers to the acquiring company, and *nonself* refers to the takeover target company. The *self* can have many strategies for selection of *nonself* that create such additional value. One seminal strategy [33] could be to search for a *nonself* that does not closely represent it or is not identical to it. In other words, the *self* would be much better off if we could identify its best compatible target among the nonselfs, whose values of financial variables are not close to its own, say outlier values. Extreme outlier value of the *self* can be found by setting the value of **n** at a high value at which $n\delta^i_{x_j}$ is marginally less than $\mu^i_{x_j}$. However on the converse, if the *self* wants its best compatible target to be similar to it, one might set a low value of **n**, at which the difference between $n\delta^i_{x_j}$ and $\mu^i_{x_j}$ is large.

Step 4a and 4b: In the next step we match the feature space vector of the *self* **A** against its detector vector space. In step 4a, we apply the rule that for any value of a feature variable xij that falls within the range of $[(\mu^i_{x_j} - n\delta^i_{x_j} - C), (\mu^i_{x_j} - n\delta^i_{x_j} + C)]$ is rejected, and the rest are accepted. Step 4b is shown by Fig. 4 where the matching/overlapping of the feature space vector and the detector space vector produce the space vector of the Accepted detectors space vector.

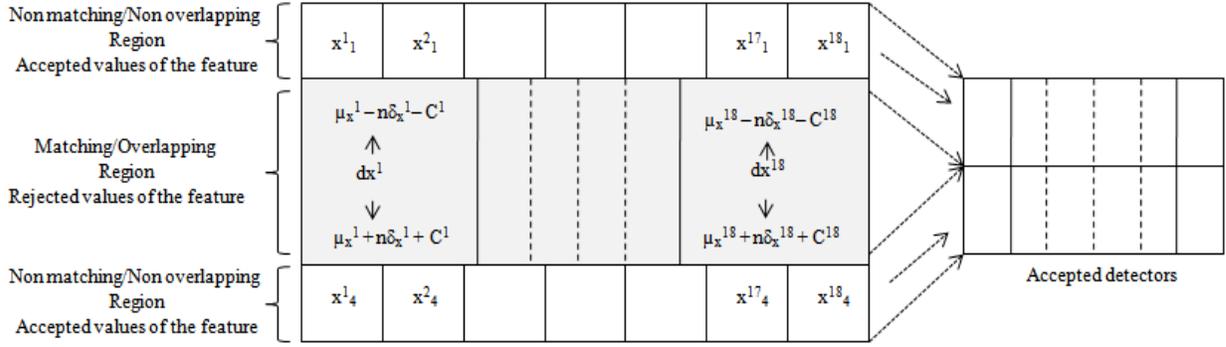

Fig. 4: Derivation of Accepted detectors space vector, in the working example.

**Step 5a and 5b – Monitoring phase**: In these last steps we aim at finding the best compatible future takeover target(s) among the *nonself* potential takeover targets. As noted above the working example, they are: {**b, c, d, e, f, g, h**}. So one wants what is most similar in characteristic to the outlier values of the *self* **A** (or most dissimilar to **A**). Let us denote the value of a feature variable of a *nonself* company by $nx^{ik}_i$, where **nx** denotes the value of a feature variable **i**, **j** denotes the year, and **k** denotes the *nonself* company, i.e., **k** = {**b, c, d, e, f, g, h**}. A tabulation of these $nx^{ik}_j$ values gives us the feature space vector of the *nonself* companies (**Fig. 5**).

In the final step we compare each of the $x^i_j$ values of the *self* firms against the $nx^{ik}_j$, for the same value of **i**, i.e. a comparison is made between the same feature variables of the *self* and the *nonself*s others. In the working example that would be seven potential takeover targets across four years of data. This comparison (interchangeably referred to the AIS literature as "monitoring" [3]) can be done through a number of ways, namely, by measuring the Euclidian distance or the cosine similarity of angles between the variables. Finally, the value of the average (for all features) of the distances provides the best compatible target from among the potential takeover *nonself* targets.

| $nx^{1b}_1$ | .... | .... | .... | .... | .... | .... | $nx^{18b}_1$ |
|---|---|---|---|---|---|---|---|
| | | | | | | | |
| $nx^{1b}_4$ | .... | .... | .... | .... | .... | .... | $nx^{18}_4$ |
| . | .... | .... | .... | .... | .... | .... | . |
| . | .... | .... | .... | .... | .... | .... | . |
| . | .... | .... | .... | .... | .... | .... | . |
| $nx^{1h}_1$ | .... | .... | .... | .... | .... | .... | $nx^{18h}_1$ |
| | | | | | | | |
| $nx^{1h}_{18}$ | .... | .... | .... | .... | .... | .... | $nx^{18h}_4$ |

Fig.5: Feature space vector of the *nonself* companies {**b,c,d,e,f,g,h**}

## V. CASE STUDY

Problem environment: The selected case study is based on real financial data from the Indian banking sector. The data is obtained from the publicly available database [34] of the Reserve Bank of India which acts as a regulatory authority of the Indian banking system. State Bank of India (SBI) is one of the largest banks in India. Prior to 2008, SBI had seven associate banks – State Bank of Saurashtra (BS), State Bank of Indore (BI), State Bank of Mysore (BM), State Bank of Patiala (BP), State Bank of Hyderabad (BH), State Bank of Travancore (BT), and State Bank of Bikaner and Jaipur (BB). Industry reports [35] [36] indicated that SBI was interested in strengthening its position among its Indian and global rivals by acquiring its associate banks. The bank believed that such acquisitions (of its associate banks) would increase its efficiency of scale, better utilize capital and scarce human resources, and better exploit brand equity [37]. It was under such a scenario that SBI acquired one of its smallest associate BS [36], in mid-2008.

Problem statement: In the present case we look to the period prior to the mid-2008 when SBI was searching among its seven associate banks (potential takeover targets) to find its best compatible target. Although we know that it finally acquired BS, we would like to come to this conclusion through the above proposed theoretical methodology. If we are able to do so, we might be in a position to recommend NSA as a possible approach that can be applied for target identification in M&A scenarios. Based on the fact that BS was the smallest among the associate banks and SBI was the largest among the Indian banks, we assume that SBI's strategy was to acquire one bank among its associate banks that was very dissimilar in characteristics to it. However, most importantly, it is to be noted that unlike the common financial management practice of Ratio Analysis[2], we do not measure similarity/dissimilarity

---
[2] In Ratio Analysis, ratios are calculated from the various financial information of a firm's balance sheets to quantitative measure it's financial performance over a time period. Similar ratios are often compared across

between the banks as comparison between their respective financial variables. Instead we do measure similarity/dissimilarity as distances (Euclidean distance) and angles (cosine similarity) between the considered financial variables.

Application of proposed Methodology: In this section we discuss how the theoretical aspects of Section IV can be applied in practice. As evident from the above two paragraph, here the acquirer or the *self* firm is SBI; and the potential takeover targets or *nonself* are the seven associate banks. Since the acquisition process starts after July 2008, the period of analysis is considered as four years from 2005 to 2008. Based upon prior literature [10] [16] [17] [20] [21] [38], a broad list of 18 financial variables are chosen as features to describe the *self* and *nonself* banks (check features in Table1).

In a first preprocessing step, the data is normalized using the formula $(x - x_{min})/(x_{max} - x_{min})$ in order to map them on a common and comparable scale. Here **x** refers to the value of a particular feature of a bank for each year. $x_{min}$ and $x_{max}$ refer to the minimum and the maximum value of the same feature across the four years considered for the analysis. Since the considered data is of four years and 18 variables are considered, the feature space of the *self* (SBI) consists of 18x4 dimensions. Also for the seven associate (potential takeover targets) banks for which 18 features are considered for 4 years, the total feature space of the *nonself* banks is of 7 x 18 x 4 dimensions. The next step is to find the detector space of the *self* (SBI). At this point, we recall that SBI's strategy is to find a takeover target (among the *nonself* banks) that is most dissimilar to it. In order to find such dissimilarity, it is essential to find that part of the feature space of the *self* (SBI) that characterizes itself – theoretically this is the detector space of the *self* (SBI). We understand that - one way of characterizing the *self* is to use for each feature, a range with lower limit [**μ– (n)*δ– c**] and higher limit [**μ + (n)*δ+ c**]. Here **μ** and **δ** are the mean and standard deviation of each feature over the four years period, and **c** denotes the change. Values of all features that fall within the range are characteristic of SBI. Empirically this is the detector space of SBI. We empirically found that setting **n** to 0.45 achieved the best results. Also uncertainties in external and internal environmental factors affect the functioning of any business and the M&A decisions of any acquirer [39].

In our present analysis we incorporate such uncertainty by stipulating change = **u**[-1,1] times the mean of average rate of growth[3] of a feature across the four years, where **u**[-1,1] is a random number from a uniform distribution in the range [-1,1] that includes some uncertainty. Subsequently we superimpose the detector space of SBI on its feature space to generate the accepted detectors space. Conceptually we can visualize the accepted detector space of SBI to be the feature space of SBI minus the superimposed part between its detector space and its feature space. In practice we generate the accepted detectors space of SBI as such: for each feature, all values of its feature space that fall within the range of the lower and the upper limit of the detector space is set to zero. The remaining values of the features that fall beyond the range are kept.

Finally, we compare the accepted detectors space of SBI with the feature space of the seven *nonself* banks in order to quantitatively measure the similarity/dissimilarity between the acquirer and the potential takeover targets, respectively. Such a comparison is done through two measures – the Euclidean distance measurement, and angle measurement based on cosine similarity. From 1000 repeated trials**,** tables 2 and 3 shows the result for the two measures. It is to be noted that for both the tables, rank 1 and rank 7 denote the largest and the smallest Euclidean distance and angle respectively, between SBI and the takeover targets. We use 1000 repeated trials in which the **u** can take any random value – (minus 1, 0, or plus 1), and the uncertainty due to the environmental factors is handled through this way. We believe that stable results can be obtained under large number of trials (1000).

Comments upon Obtained Results: The results from Table 2 suggest that for these 1000 repeated trials, NSA is able to correctly predict for 85% of cases that BS is farthest in distance (note that rank 1 refers to the largest Euclidean distance) and hence most dissimilar to SBI. Similar conclusions can be drawn from Table 3 that shows that in 80.2% of the cases, NSA is able to correctly predict that BS has the farthest divergence in angle to SBI. Since the above results corroborate with the actual fact (in above 80% of cases) that SBI acquired BS in the past, we might conclude that NSA has been able to correctly identify takeover target in M&A scenarios in this particular case.

We further compare the above results with the correlation coefficients between SBI and its associate banks that independently measure the degree of similarity/dissimilarity between the acquirer and the potential targets. Fig. 6 shows the correlation coefficients (between SBI and associate banks) that were calculated by averaging each feature across the four year time period. It is to be noted here that the correlation analysis is an additional support to the results obtained from NSA. This is because even though correlation coefficients give idea about the strength of similarity/dissimilarity between the acquirer and the potential targets, it is static in nature. On the contrary, the use of **u** in our analysis incorporates change due to the dynamicity in the external environment.

BS has the lowest correlation coefficient, it is most dissimilar to SBI. Similarly BI (with the highest correlation coefficient) is the most similar to SBI. So we can correctly predict that BS is to be the first preference as acquisition target

---

firms to evaluate M&A decisions. For M&A purposes in the Indian banking industry , the ratios that are commonly used are Gross Profit Margin, Return on Equity, Return on Capital Employed, Debt Equity Ratio, etc [42].
[3] The average rate of growth of a feature is calculated as {(Value of the feature for the present year – Value of the feature for last year)/ Value of the feature for last year}*100

of SBI and BI to be the least preferred target at 1.9% of cases. Even though the algorithm in its present form is limited in predicting the preference ranking of BM, BP, BH, BT and BB, yet we believe that the prediction error might be due to the high dimensionality of the feature space of the associate banks associated with the sparse data [40]. It might be recalled here that the sparse space of the accepted detectors space of SBI was created when for each feature, all values of SBI's feature space that fell within the range of the lower and the upper limit of the detector space was set at zero. In comparison angle measurement based on cosine similarity correctly predicted BS as the first preference as acquisition target. It also correctly predicted BM as the second preferred acquisition target after BS. Also we believe the problem of inaccuracy of prediction is created by the inherent assumption of cosine similarity – that the features (financial variables) in the feature space of SBI are independent and that the space is orthogonal [41]. However in actual practice the financial variables are complexly related to one another.

TABLE II. RANKING BASED ON DISTANCE MEASUREMENT.

| Rank | BS | BI | BM | BP | BH | BT | BB |
|---|---|---|---|---|---|---|---|
| 1 | 850 | 19 | 0 | 0 | 0 | 0 | 131 |
| 2 | 97 | 131 | 40 | 0 | 15 | 0 | 717 |
| 3 | 43 | 410 | 137 | 0 | 280 | 0 | 130 |
| 4 | 10 | 308 | 276 | 0 | 375 | 10 | 21 |
| 5 | 0 | 123 | 487 | 23 | 325 | 41 | 1 |
| 6 | 0 | 9 | 40 | 718 | 5 | 228 | 0 |
| 7 | 0 | 0 | 20 | 259 | 0 | 721 | 0 |

TABLE III. RANKING BASED ON ANGLE MEASUREMENT.

| Rank | BS | BI | BM | BP | BH | BT | BB |
|---|---|---|---|---|---|---|---|
| 1 | 802 | 57 | 62 | 0 | 47 | 0 | 32 |
| 2 | 86 | 96 | 634 | 0 | 114 | 0 | 70 |
| 3 | 66 | 404 | 108 | 0 | 236 | 0 | 186 |
| 4 | 25 | 194 | 108 | 1 | 260 | 6 | 406 |
| 5 | 18 | 236 | 84 | 6 | 328 | 26 | 300 |
| 6 | 1 | 11 | 4 | 342 | 15 | 621 | 6 |
| 7 | 2 | 0 | 0 | 651 | 0 | 347 | 0 |

VI. CONCLUSION

In this paper, we have contributed to the present literature of takeover target prediction in M&A scenarios by proposing a new approach and model where NSA can be applied to identify takeover targets by acquiring firms that want to carry out their first acquisition. We believe that there exists a considerable need in the market for such a model among these novice acquiring firms.

Although our current conclusions are based only on one case study, it looks very promising, especially based on other current verifications of our methodology. The next step is to further implement the proposed model on various other situations to substantiate our results and in order to be able to make a more general applicable statement about the applicability of NSA for takeover target prediction. The algorithm is being updated by the incorporation of cybernetic arches in two parts: (i) refining the determination of features of *self* and (ii) refining the determination of features of *non-self*. Another important modification though is to equip our current approach to deal with correlated features.

REFERENCES

[1] Bloomberg. (2013, Feb.) Bloomberg Web site. [Online]. http://www.bloomberg.com/professional/files/2013/01/MA-Financial-Ranking-League-Tables-2012.pdf
[2] J. Kinnunen and M. Collan, "Supporting the Screening of Corporate Acquisition Targets," in Proceedings of the 42nd Hawaii International Conference on System Sciences, Hawaii, 2009, pp. 1-8.
[3] D. Dasgupta and L. F. Nino, Immunological Computation - Theory and Applications, 1st ed. Boca Raton: Auerbach Publications, 2009.
[4] U. Rege, "Accounting Ratios to Locate Take-over Targets.," Journal of Business Finance and Accounting, pp. 301-11, 1984.
[5] D. L. Stevens, "Financial Characteristics of Merged Firms: A Multivariate Analysis," Journal of Financial and Quantitative Analysis, vol. 8, pp. 149-58, 1973.
[6] P. Barnes, "Can takeover targets be identified by statistical techniques? Some UK evidence," The Statistician, vol. 47, no. 4, pp. 573-91, 1998.
[7] R. S. Harris, J. F. Stewart, D. K. Guilkey, and W. T. Carleton, "Characteristics of Acquired Firms: Fixed and Random Coefficients Probit Analysis," Southern Economic Journal, vol. 49, no. 1, pp. 164-184, 1982.
[8] D. Castagna and Z. P. Matolcsy, "Accounting Ratios and Models of Takeover Target Screens: Some Emperical Evidence," Australian Journal of Management, vol. 10, no. 1, pp. 1-15, 1985.
[9] Meador, P. Church, and L. Rayburn, "Development of Prediction Models for Horizontal and Vertical Mergers," Journal of Financial and Strategic Decisions, vol. 9, no. 1, pp. 11-23, 1996.

[10] Tsagkanos, A. Georgopoulos, and C. Siriopoulos, "Predicting Takeover Targets: New Evidence from a Small Open Economy," International Research Journal of Finance and Economics, vol. 4, pp. 183-92, 2006.

[11] J. K. Dietrich and E. Sorensen, "An application of logit analysis to prediction of merger targets," Journal of Business Research, vol. 12, no. 3, pp. 393-402, 1984.

[12] K. G. Palepu, "Predicting Takeover Targets: A Methodological and Emperical Analysis," Journal of Accounting and Economics, pp. 3-35, 1986.

[13] R. G. Powell, "Takeover Prediction Models and Portfolio Strategies: A Multinomial Approach," Multinational Finance Journal, vol. 8, pp. 35-72, 2004.

[14] A. A. Blake. (2007) The Predicton of Australian Takeover Targets: A Binomial and Multinomial Logit Analysis. [Online]. http://ses.library.usyd.edu.au/bitstream/2123/2215/1/Blake.pdf

[15] G. Brar, D. Gimoundis, and M. Liodakis, "Predicting European Takeover Targets," European Financial Management, vol. 15, no. 2, p. 430–450, 2009.

[16] P. Barnes, "The identification of U.K takeover targets using published historical cost accounting data: Some emperical evidence comparing logit and linear discriminant analysis and raw financial ratios with industry-relative ratios," International Review of Financial Analysis, vol. 9, no. 2, pp. 147-62, 2000.

[17] H. Liu and Y. Hu, "An Application of Hopfield Neural Network in Target Selection of Merger and Acquisitions," in International Conference on Business Intelligence and Financial Engineering, Beijing, 2009, pp. 34-37.

[18] J. J. Cheh, R. S. Weinberg, and K. C. Yook, "An Application of an Artificial Neural Network Investment System to Predict Takeover Targets," The Journal of Applied Business Research, vol. 15, no. 4, pp. 33-45, 1999.

[19] S. Paul, A. Janecek, F. d. L. N. Buarque, and T. Marwala, "[In Press] A new Approach for Suggesting Takeover Targets based on Computational Intelligence and Information Retrieval Methods: a Case Study from the Indian Software Industry," in Mathematics of Uncertainty Modeling in the Analysis of Engineering and Science Problems, S. Chakraverty, Ed. Pennsylvania, United States of America: IGI Global, 2013.

[20] L. Hongjiu, H. Yanrong, and F. Shufen, "Study of Merger and Acquisition Target Selection Based on Self-Organizing Mapping Neural Network," in IEEE International Conference on Management of Innovation and Technology, vol. 2, Singapore, 2006, pp. 1035-38.

[21] E. Little, R. Hickey, and A. Brabazon, "Identifying Merger and Takeover Targets Using a Self-Organizing Map," in World Congress in Computer Science Computer Engineering and Applied Computing, Las Vegas, 2012. [Online]. <a href="http://ww1.ucmss.com/books/LFS/CSREA2006/ICA3076.pdf">http://ww1.ucmss.com/books/LFS/CSREA2006/ICA3076.pdf</a>

[22] R. L. King, S. H. Russ, A. B. Lambert, and D. S. Reese, "An artificial immune system model for intelligent agents," Future Generation Computer Systems, vol. 17, pp. 335-43, 2001.

[23] U. Aickelin and D. Dasgupta. arXiv - Cornell University Library. [Online]. http://arxiv.org/ftp/arxiv/papers/0910/0910.4899.pdf

[24] J. Timmis, A. Hone, T. Stibor, and E. Clark, "Theoretical advances in artificial immune systems," Theoretical Computer Science, vol. 403, pp. 11-32, 2008.

[25] L. N. Castro and J. Timmis, Artificial Immune Systems: A New Computational Intelligence Approach, 1st ed. Berlin, Germany: Springer, 2002.

[26] Prakash and S. G. Deshmukh, " A multi-criteria customer allocation problem in supply chain environment: An artificial immune system with fuzzy logic controller based approach," Expert Systems with Applications, vol. 38, pp. 3199-3208, 2011.

[27] Y. Chiu, I. T. Kuo, and C. H. Lin, " Applying artificial immune system and ant algorithm in air-conditioner market segmentation," Expert Systems with Applications, vol. 36, pp. 4437-4442, 2009.

[28] Brabazon, M. O'Neill, and J. Dempsey, "An Introduction to Evolutionary Computation in Finance," IEEE Computational Intelligence Magazine, vol. 3, no. 4, pp. 42-55, 2008.

[29] Fyfe and L. Jain, "Teams of intelligent agents which learn using artificial immune systems," Journal of Network and Computer Applications, vol. 29, pp. 147-159, 2006.

[30] M. Danziger, M. Lacerda, and F. B. d. Lima Neto, "A Hybrid Approach for IEEE 802.11 Intrusion Detection Basedon AIS, MAS and Naïve Bayes," Internationa lJournal of Computer Information Systems and Industrial Management, vol. 3, p. 193–201, 2011.

[31] M. J. Epstein, "The determinants and evaluation of merger success," Business Horizons, pp. 37-46, 2005.

[32] R. B. Mason, "The external environment's effect on management and strategy," Management Decision, vol. 45, no. 1, pp. 10-28, 2007.

[33] H. Liu and W. Ma, "Price Prediction of Target of Mergers and Acquisitions Based on Genetic-algotihm BP Neural Network," in IEEM 2009, Hong Kong, 2009, pp. 787-91.

[34] R. B. of. India. Database on Indian Economy. [Online]. http://dbie.rbi.org.in/DBIE/dbie.rbi?site=publications#!4

[35] G. Press Information Bureau. (2008, Jul.) Merger of State Bank of Saurashtra with State Bank of India. [Online]. http://pib.nic.in/newsite/erelease.aspx?relid=40587

[36] moneycontrol. (2008, Jul.) Cabinet okays SBI, State Bank of Saurashtra merger. [Online]. http://www.moneycontrol.com/news/business/cabinet-okays-sbi-state-banksaurashtra-merger-_348724.html

[37] R. N. Pradeep. (2012, Dec.) Public sector banks, consolidate now. [Online]. http://www.thehindubusinessline.com/opinion/public-sector-banks-consolidate-now/article4210158.ece

[38] W. G. Kim and A. Arbel, "Predicting merger targets of hospitality firms (a Logit model)," Hospitality Management, vol. 17, pp. 303-18, 1998.

[39] U. Steger and C. Kummer. (2007) IMD. [Online]. https://www.imd.org/research/publications/upload/Steger_Kummer_WP_2007_11.pdf

[40] M. E. Houle, H.-P. Kriegel, P. Kroger, E. Schubert, and A. Zimek, "Can Shared-Neighbor Distances Defeat the Curse of Dimentionaity," in 22nd International Conference, SSDBM, Heidelberg, 2010, pp. 482-500.

[41] N. Liu, et al. (2004, Nov.) learning Similarity Measures in Non-orthogonal Space. [Online]. www.bradblock.com.s3-website-us-west-1.amazonaws.com/Learning_Similarity_Measures_in_Non-orthogonal_Space.pdf

[42] Khan, "Merger and Acquisitions (M&As) in the Indian Banking Sector in post Liberalization Regime," International Journal of Contemporary Business Studies, vol. 2, no. 11, pp. 31-45, Nov. 2011.